\documentclass{article}
\usepackage{arxiv}
\usepackage[utf8]{inputenc} % allow utf-8 input
\usepackage[T1]{fontenc}    % use 8-bit T1 fonts
\usepackage{hyperref}       % hyperlinks
\usepackage{url}            % simple URL typesetting
\usepackage{booktabs}       % professional-quality tables
\usepackage{amsfonts}       % blackboard math symbols
\usepackage{amsmath}       % blackboard math symbols
\usepackage{nicefrac}       % compact symbols for 1/2, etc.
\usepackage{microtype}      % microtypography
\usepackage{lipsum}		% Can be removed after putting your text content
\usepackage{graphicx}
\usepackage{natbib}
\usepackage{doi}

\title{Learning to generate feasible graphs using graph grammars}

\author{
	{Stefan Mautner, Rolf Backofen}\\
	Department of Computer Science\\
	University of Freiburg\\
	\texttt{\{mautner,backofen\}@informatik.uni-freiburg.de}\\
	\And
	{Fabrizio Costa} \\
	Department of Computer Science\\
	University of Exeter\\
	\texttt{F.Costa@exeter.ac.uk}
}

\renewcommand{\headeright}{}
\renewcommand{\undertitle}{}
\renewcommand{\shorttitle}{}

\hypersetup{
pdftitle={Learning to generate feasible graphs using graph grammars},
pdfsubject={graph-grammars},
pdfauthor={Stefan Mautner, Fabrizio Costa},
pdfkeywords={Graph grammars, pareto, ranking SVM},
}

\begin{document}

\maketitle

\begin{abstract}

Generative methods for graphs need to be sufficiently flexible to model complex
dependencies between sets of nodes. At the same time, the generated graphs need
to satisfy domain-dependent feasibility conditions, that is, they should not
violate certain constraints that would make their interpretation impossible
within the given application domain (e.g. a molecular graph where an atom has a
very large number of chemical bonds). Crucially, constraints can involve not
only local but also long-range dependencies: for example, the maximal length of a
cycle can be bounded.

Currently, a large class of generative approaches for graphs, such as methods
based on artificial neural networks, is based on message passing schemes. These
approaches suffer from information 'dilution' issues that severely limit the
maximal range of the dependencies that can be modeled. To address this problem,
we propose a generative approach based on the notion of graph grammars. The key
novel idea is to introduce a domain-dependent coarsening procedure to provide
short-cuts for long-range dependencies.

We show the effectiveness of our proposal in two domains: 1) small drugs and 2)
RNA secondary structures. In the first case, we compare the quality of the
generated molecular graphs via the Molecular Sets (MOSES) benchmark suite,
which evaluates the distance between generated and real molecules, their lipophilicity, synthesizability, and drug-likeness. In the second case, we show
that the approach can generate very large graphs (with hundreds of nodes) that
are accepted as valid examples for a desired RNA family by the ``Infernal''
covariance model, a state-of-the-art RNA classifier.

Our implementation is \textbf{available} on github:
github.com/fabriziocosta/GraphLearn

\end{abstract}

\section{Introduction}

Generative methods for graphs are receiving increased attention due to the
variety and importance of application domains that can be naturally modeled
using this formalism. Operating on discrete structures however often incurs
computational problems due to the combinatorial complexity explosion. One
popular domain where these problems are manageable is organic chemistry, where
data is plentiful and the size of the discrete label's alphabet and of the
overall molecular graph are small \citep{MOSES}. However, larger molecules, such
as RNAs or proteins, exhibit an additional source of difficulties, i.e. the
presence of complex dependencies between their different constituent parts.

In \cite{costa16} the problem of generating elements of a
structured domain was framed as the
equivalent problem of sampling from a corresponding underlying probability
distribution defined over a (learned) class of structures. Specifically, they
employ a context-sensitive grammar to accurately model complex dependencies
between different parts of an instance. The authors acknowledge that an
approach based exclusively on a grammar is not sufficient since the number of
proposed graphs grows exponentially with the number of production rules in the
grammar. To address the issue, it was proposed to use a Metropolis
Hastings (MH) Markov Chain Monte Carlo (MCMC) method, where the sampling problem is reduced to the easier task of {\em simulation}. 
They use context-sensitive graph grammar to inform the MH proposal distribution, but
also introduce a probability density estimator to define the MH acceptance
procedure. This allows us to deal separately with local and global constraints:
the locally context-sensitive graph grammar is used for the local constraints
and the regularized statistical model is used for the global or long range
constraints. The two approaches complement each other: the grammar is a
flexible non-parametric approach that can model complex dependencies between
vertices that are within a short path distance from each other; the
statistical model, instead, can employ the bias derived from the particular
functional family (linear) and the type of regularization used (a penalty over
the $\ell_2$ norm of the parameters) to generalize long range dependencies to
similar cases. This approach is therefore adequate when the underlying concept
exhibit local dependencies that are more complex than long range ones.
Unfortunately in some application domains instances can exhibit complex long
range dependencies between their different constituent parts. This is the case
for RNA polymers, long sequences of atomic entities (nucleotides) that self
interact, establishing pairwise hydrogen bonds that can typically span the entire
length of the sequence, where sequences can range from hundreds to thousands of atomic parts. 

A different way to view the issue of long range dependencies is that of the
selection of an appropriate scale of representation at which to work and reason.
Certain application domains exhibit natural encodings, i.e. instances are
encoded as graphs where nodes represent specific entities, such as nucleotides
in the case of RNA sequences. However, it is known that a more effective
functional description of RNA polymers can be obtained in terms of structural
components such as {\em stems} (stretches of consecutive paired nucleotides) and
{\em loops} (stretches of consecutive unpaired nucleotides). Under this view,
dependencies that are local at the coarser scales correspond to longer range
dependencies at the original scale.

A constructive system suitable for these domains needs to be able to
adequately model complex long range dependencies and is more effective if it
operates at a convenient coarser scale rather than that of the individual
units. Here we tackle all these issues extending the approach presented in
\cite{costa16} with two key ideas: 1) we allow a user/domain defined graph
coarsening procedure, and 2) we allow domain specific optimization procedures
to ensure that generated instances are always viable.

\section{Method}

\cite{costa16} presented an approach to sample graphs from an empirical
probability distribution using a variant of the Metropolis Hastings (MH)
algorithm \citep{metropolis1953}. The MH approach decomposes the
transition probability of an underlying  Markov process in two factors, 1) the
proposal  and 2) the acceptance-rejection probability distributions. The
algorithm starts from a seed element and iteratively proposes new instances
that are stochastically accepted if they fit appropriate
requirements. The key element for an efficient MH design is building the
proposal distribution in such a way that the generated elements will not be
rejected too often. To do so Costa suggests to use a graph grammar and to infer its
rules from the available data using grammatical inference techniques upgraded
to structured domains (i.e. domains where instances are naturally represented
as labeled graphs).
A graph grammar \citet{rozenberg1999handbook} is a formal framework used to specify rules for graph transformation. It involves a set of productions denoted as $(M, D, E)$, where $M$ represents the 'mother' subgraph that is to be replaced, $D$ is the 'daughter' subgraph that replaces $M$, and $E$ is the embedding mechanism that governs how $D$ is incorporated into the overall graph structure. This formalism allows for the systematic generation of complex graph structures through the iterative application of these transformation rules.
\cite{costa16} introduced an efficient graph grammar based on the concept of
{\em distributional semantics}
\citep{harris1954distributional,harris1968mathematical} and on the {\em
substitutability principle} \citep{Clark:2007}. The notion of mother and
daughter graphs are specialized here as neighborhood graphs that share the
``outer'' context, and the embedding mechanism is simplified by imposing an
exact match of the context (see Fig. \ref{allcips}). The context is called {\em
interface} and its size is parametrized by a discrete value $T$ ($T=2$ means
that the ``thickness'' of the context comprises two nodes), while the inner part
is called {\em core} and its size is parametrized by the radius $R$. The
embedding mechanism can now replace a core-interface pair (CIP) with another,
provided that the interfaces are isomorphic. Note that when the radii have
different values, the overall size of the generated (daughter) graph is reduced
or increased with respect to the size of the seed (mother) graph. One of the
advantages of this type of grammar is that the set of all available production
rules can be induced efficiently from a set of graphs (see \citep{costa16} for
details). More precisely, a {\em core} graph $C^v_R(G)$, is a neighborhood graph
of $G$ of radius $R$ rooted in $v$. An interface graph, denoted $I^v_{R,T}(G)$
is the difference graph of two neighborhood graphs of $G$ with the same root in
$v$ and radii $R$ and $R+T$. In Fig. \ref{allcips} the core $C_{R}^v(G)$ (dark)
is determined by vertices at maximal distance $R$ ($R=0$ in the figure) from a
chosen root vertex $v$. The interface $I_{R,T}^v(G)$ with \emph{thickness} $T=2$
is in a lighter shade. Given a new core-interface pair (CIP) with matching
interface, the substitution can take place to yield the replacement of a carbon
with a nitrogen atom.

\subsection{Contribution}

In the context of RNA polymers, two key challenges emerge: (1) {\em resolution} and (2) {\em viability}. First, the level of resolution impacts efficiency, as replacing only a few nodes per iteration is insufficient for structures with hundreds of nodes. Second, viability pertains to the structural integrity of the generated graphs—small alterations can lead to invalid structures.  It is known that for RNAs a more effective description is
obtained considering larger structural components such as {\em stems}  and {\em
loops} and not working at the resolution scale of single nucleotides. As for the
second issue, it is known that the function of RNA polymers depends on their
global structure (i.e. the set of pairs of interacting nucleotides), which can
significantly change when even a single nucleotide is altered. To deal with
these issues we propose two enhancements: 1) a grammar coarsening procedure and
2) a constraint integration procedure.
The following section elaborates on the details of this approach, demonstrating its effectiveness across two case studies.

\begin{figure}[ht]
      \centering
        \includegraphics[width=0.7\linewidth]{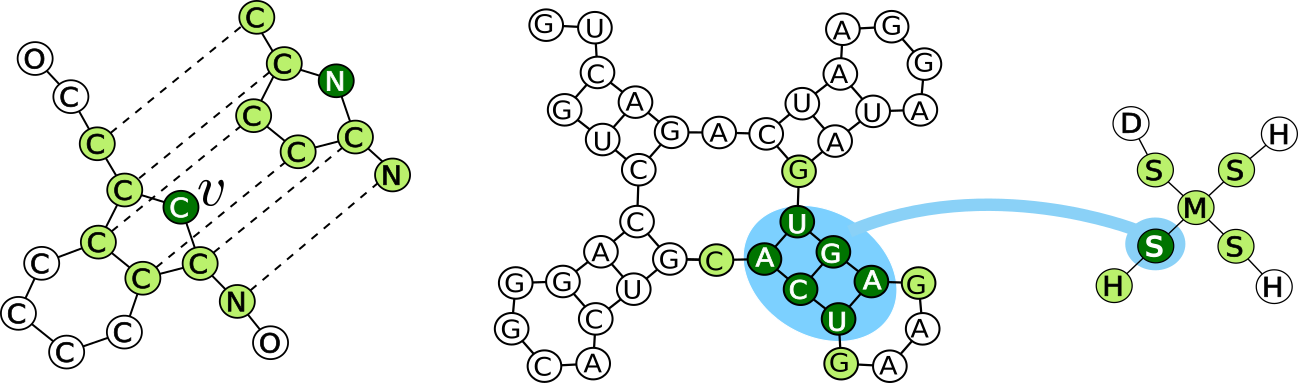}
      \caption{
      A depiction of core and interface subgraphs. Dark shading indicates the core subgraph, while light shading represents the interface. The diagram on the left shows a molecular graph with its core substitution, while the middle and right diagrams depict RNA encodings at different resolutions (nucleotide-level and structural-element-level, respectively). At nucleotide-level nodes are labeled with nucleotides codes ACGU, while at structural level nodes are labeled as H)airpin, M)ultiloop, S)tem, D)angling end.}
      \label{allcips}
\end{figure}

\textbf{Grammar coarsening.} The idea is to allow users to specify
a coarsening procedure via the notion of {\em edge
contraction}:
an operation which removes an edge from a graph while simultaneously merging
the two vertices that it previously joined. In addition, we allow for the more
general notion of {\em vertex contraction}, which removes the restriction that
the contraction must occur over vertices sharing an incident edge. Both
procedures are defined using a node attribute $cid \in \mathbb{N}$ called {\em
contraction-identifier} and contracting all vertices that share the same $cid$.
We propose to use the contraction operation as a flexible way to
transform a graph to its coarser version, $G \mapsto G'$. See Fig. \ref{allcips}
for an example of coarsening. Cores and interfaces
can now be defined exploiting the contracted graph. Starting from a CIP on the
coarse graph, we define the core as the subgraph induced by the vertices of
the original graph that have been contracted to vertices of the core of radius
$R$ in the coarse graph, $C_R^v(G',G)$.  The new interface graph
$I_{R,B}^v(G',G)$ is defined as the Cartesian product of the graph induced by
the nodes adjacent to the core nodes in $G$ at maximal distance $B$ (the
thickness in the base graph) and the interface graph on the coarse graph. In
words, we require that both the interface at base level and at coarse level
match for a core swap to take place. An RNA sequence is a sequence over
the nucleotides
$A,C,G$ and $U$, these nuceleotides may form hydrogen bonds, forming a structure:
$A$ with $U$ and $C$ with $G$. In Fig. \ref{allcips} we depict an RNA
polymer graph at the nucleotide level (center) and its coarse version (right)
where the contraction was informed by the notion of structural components such
as stems, hairpin loops, and multi-loops. Note that a core at coarse
level (in dark) made only of a single node corresponds to multiple nodes at
base level. Moreover, while the interface at base level requires only the
presence of few nucleotides, these have to belong to a context defined at the
coarse level that can span a much larger fraction of the instance and can be
viewed as a global localization indicator.

\textbf{Constraint integration.} It should be possible to easily make use of
specific feasibility constraints when they are available for a given domain.
For RNAs the relation between sequence and structure is known to be governed
by thermodynamical forces that seek to minimize the amount of free energy of
the molecule. The structure of a given sequence can be computed using
sophisticated dynamic programming optimization algorithms. To integrate these
constraints in the constructive protocol we let the user specify a
transformation function which maps candidate instances to feasible instances.
In our case, the transformation takes a graph representing an RNA structure
constructed by the proposed procedure, removes the pairwise nucleotide bonds, and
recomputes them as the solution of a given folding algorithm. This
transformation ensures that we are always computing the acceptance probability
on viable RNA structures.

\begin{figure}[ht] \centering
        \includegraphics[width=0.35\linewidth]{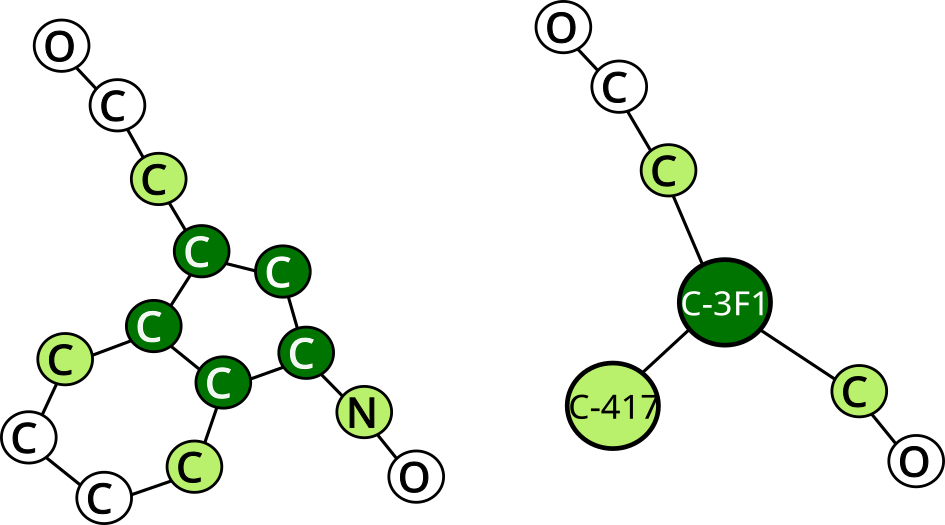} \caption{ A
        molecule and its coarsened version $R=0$ (dark green), $B=1$ (left, lt.
        green), $T=1$ (right, lt. green). Local constraints are satisfied by the
        original graph on the left while longer ranging constraints
        (specifically the fact that the carbon atoms are not only present but
        form a cycle) is modeled in the coarsened version (right). The
        coarsening method labeled contracted cycles with the hash of the
        associated subgraph. } \label{chemcoarse} \end{figure}

\textbf{Chemical compounds case.} In addition to RNA polymers we consider also
small drugs. One type of long range dependency for organic molecules is the
membership to ring structures (see Fig. \ref{chemcoarse}). In this case, the
context matching procedure is sufficient to guarantee that the CIP replacement
operation will almost always generate feasible molecular structures, which
eliminates the necessity of a constraint integration procedure.

\section{Empirical evaluation}

\begin{figure}[ht]
      \centering
        \includegraphics[width=0.7\linewidth]{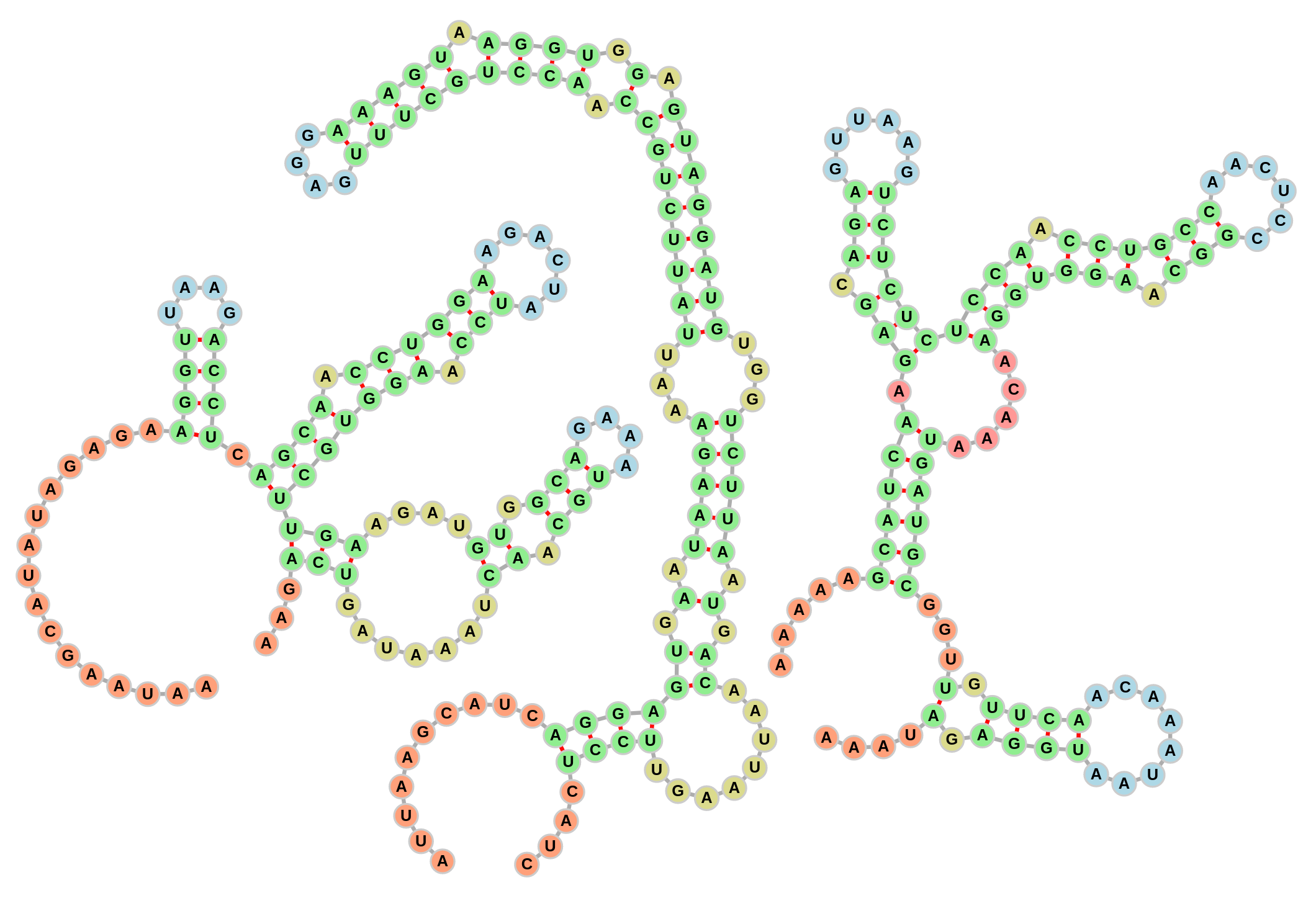}
      \caption{Examples of the rf01725 family of RNA, exhibiting a large variety of structural configurations.}
      \label{RF01725}
\end{figure}

Ribonucleic acid (RNA) consists of a chain of nucleotides connected via a ribose backbone.
RNA polymers cover essential biological roles ranging from coding, decoding,
regulation and expression of genes.
Specifially in this work we are looking at non coding RNAs which are not translated into proteins but
self interact, forming a structure which is indicative of its function.
The RFam database \citep{rfam} groups
known RNA sequences in functionally and evolutionarily related {\em families}. To
empirically investigate the performance of the proposed constructive approach
we selected the SAM-I/IV variant riboswitch (RF01725) and SAM riboswitch (S
box leader) (RF00162), which are families that exhibit a rich structure (e.g.
that do not consist only of a single hairpin element)  and an average sequence length of 100.
We showcase graphs from RF01725 in Fig.\ref{RF01725}. The general aim is to
synthesize functionally equivalent but novel sequences, a task with important
biomedical applications.

When working with RNAs, the feasibility constraints are related to the
structure determination. To improve accuracy we use a two step approach:
given a candidate sequence we 1) identify $k$ nearest neighbors among the
available training instances, then 2) we align the $k+1$ sequences using the
\emph{MUSCLE} (MUltiple Sequence Comparison by Log- Expectation) program
\citep{muscle} and compute the consensus structure using the \emph{RNAalifold}
program \citep{rnaalifold}. This procedure identifies
the most representative structure in the ensemble of possible suboptimal
solutions. Here we use four neighbors. 

\begin{figure}[ht]
      \centering
  \begin{minipage}[b]{0.47\textwidth}
    \includegraphics[width=\textwidth]{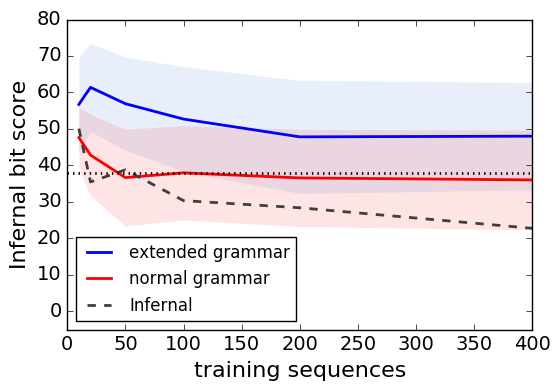}
    \caption{
    Evaluation of generated RNAs.
    The Infernal bit score is the expert-models classification
    score.
    The horizontal line is the threshold for biological significance
    associated with this RNA family.
    % We compare our method to the baseline grammar induced graph generation
    % and Infernals own covariance based sequence generation model.
    % Estimated equivalence by Infernal. The horizontal line is the threshold for biological significance.
    }
      \label{infeval}
  \end{minipage}
  \hfill
  \begin{minipage}[b]{0.52\textwidth}
    \includegraphics[width=\textwidth]{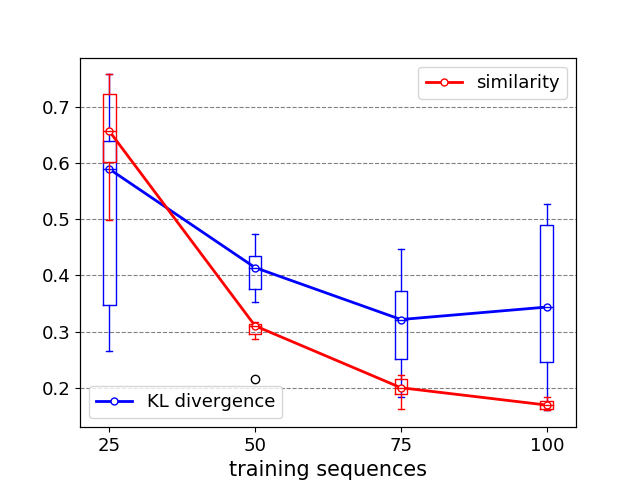}
    \caption{
    Evaluation of generated RNAs. Here we compare the
    generated sequences to the training material. 
    The generated graphs should differ from the training set, 
    yet induce the same probability density.
    }
     \label{learncurve}
  \end{minipage}
\end{figure}

In order to evaluate if the constructed sequences are functionally equivalent
to the original examples, we use a state-of-the-art classifier as the {\em
oracle} rather than resorting to a more expensive real world biological
experiment. To define the membership of a sequence to a given family, the RFam
database uses the covariance model computed by the program \emph{Infernal}
(INFERence of RNA ALignment) \citep{infernal} induced over a hand curated set of
representative sequences. In Fig.~\ref{infeval} we report the average score
achieved by the constructed sequences as the training set increases. The
horizontal line indicates the family specific threshold above which the Infernal
model accepts a sequence as a valid instance of the family. In blue we report
the results for the proposed extension and in red the results for the original
method. From the perspective of usual machine learning techniques it seems
interesting that the performance decreases when training material is added. This
is due to the sampling process. Productions are repeatedly applied to one of the
training instances. If the grammar is small, there are only few productions
applicable, resulting in a final instance that is close to training instances.
To make it comparable to our method, we used MUSCLE and RNAalifold to train a
new Infernal model and report the performance of its generated sequences
(dashed). We note that the coarsening strategy consistently improves the quality
of the results.

In order to evaluate if the generated graphs are distinct from the original
material, yet cover a similar distribution, we report an evaluation of the
similarity between these two sets. More precisely we compute two measures: 1)
the Kullback-Leibler (KL) divergence  between the probability density estimated on
the original data vs. the density estimated on the generated graphs, and 2) the
set similarity between the original and the generated data. 
For both measures we vectorize the graphs via the NSPDK graph kernel \citep{costagrave} with its
default settings.
We used the RF01725 RFam family, since
the graphs have few nodes and a relatively large structural variety.  

Specifically, we compute the \textit{symmetrized KL divergence} as:
\[
D_{KL}(P_G \parallel P_R) + D_{KL}(P_R \parallel P_G)
\]
where \( P_G \) and \( P_R \) are the probability distributions estimated from the generated and real data, respectively. To estimate these probabilities, we train two \textit{One-Class SVM} models: one on the generated data and the other on the real data.
We then use these models to evaluate the probability of a test set that consists of an equal mix of real and generated instances. The model trained on the generated data produces the probabilities \( P_G \), while the model trained on the real data generates \( P_R \).
This procedure is repeated using cross-validation, and we report the median results from these trials.

To measure the similarity between the generated and real graph sets, we employ a 
\textit{normalized kernel similarity measure}. Specifically, we compute the kernel similarity between 
each pair of elements from the set of generated instances \( G \) and the set of real instances \( R \), 
summing these values across all pairs:

\[
K(G, R) = \sum_{g \in G} \sum_{r \in R} k(g, r)
\]

where \( k(g, r) \) is the NSPDK graph kernel, which measures the similarity between an element 
\( g \in G \) (from the set of generated instances) and \( r \in R \) (from the set of real instances). 
To normalize this similarity score, we divide by the geometric mean of the self-similarity scores for 
both sets:

\[
\text{similarity}(G, R) = \frac{K(G, R)}{\sqrt{K(G, G) \cdot K(R, R)}}
\]

For each box in the box plot we collect 7 repetitions.

In Fig.~\ref{learncurve} we compute these measures as a function of the training
set size. We observe that the similarity tends to decrease as more diverse
material becomes available and that also the divergence tends to vanish,
indicating that the generated instances induce the same probability density as
the original ones, albeit being increasingly different.

\subsection{Graph generative approaches}

Deep neural network approaches to generate graphs exist, but they cannot be easily adapted to the domain of large graphs with long range dependencies.
DeepGMG \citep{deepGMG} and GraphVAE \citep{graphVAE} for example can process graphs with a maximum size of few tens of nodes.
Domain specific solutions on chemical compounds can operate on larger graphs; the MOSES \citep{MOSES} benchmark  contains a large set of graphs, these are restricted to small cycles, an alphabet size of 8 and a molecular weight of 250-350 Daltons (which results in small graphs). Several algorithms, e.g. CharRNN
\citep{smilesbasedgen} which can only operate on SMILES strings, are domain specific.

In many real world cases, the size of the available set of examples is not extremely large. To simulate this scenario we limit the train set size to 1000 instances.
We implemented the coarsening as described before and used the parameters $B=2, T=1,R=\{0,1\}$.
MOSES offers a variety of measures to assess the quality of the produced molecular graphs:
\textbf{FCD} "Frechet ChemNet Distance" a deep learning based estimator that determines how similar instances are to ``real'' molecules \citep{chemnet}; \textbf{SNN} measures how close instances are to the test set on average; \textbf{Frag and Scaf} computes the cosine similarity induced by fragment collections of test and produced sets; \textbf{IntDiv1,2} measures the internal diversity of the sampled set; \textbf{filters} looks for unstable, very reactive or toxic instances.
Furthermore they implement four algorithms for molecule sampling: VAE \citep{mosesVAE} a FAN \citep{ORGANIC} combines Generative Adversarial Networks (GANs) and reinforcement learning (RL) to work on SMILES, CharRNN emplays a Long Short-Term Memory RNN to learn the language and syntax of SMILES and AAE Adversarial autoencoder, a mix of VAEs and GANs \citep{MOSES}.

We examine the drawbacks of each approach. VAE and ORGAN produce less than 10\%
valid graphs, CharRNN and AAE produce 35\% and 36\% valid and unique graphs. The
best performing approach is CharNN with top performance in FCD, Frag and Scaf
measures. The drawback of CharNN is that it produces molecules of such small
size (5-7 atoms) that exhaustive enumeration becomes possible. AAE performs best
in SNN (this is likely due to generating graphs of the appropriate size), but low
according to the IntDiv scores(0.69/0.66, while all other generators are  in the
range 0.84 $\pm$ 0.02), suggesting that its products are very similar to the
training material (it outputs only 216 unique graphs over 604 valid ones). The
products of the non coarsened grammar perform very similar to the coarsened
version however only 79 of the 1000 generated graphs pass the filter test
(second lowest is ORGAN with 58.3\% , coarsened grammar: 87.2\%) this suggests
long range dependencies are not met, resulting in unstable molecules. The
weakness of our approach is that the generated instances tend to be large in
size which affects the SNN measure. This tendency is a byproduct of the
acceptance procedure and in future work will be addressed. In summary, our
procedure, while being domain agnostic, generates 100\% viable graphs, with
comparable performance to the other specialized algorithms (see
Tab.~\ref{mosestab}).

\begin{table}[]
\centering
\begin{tabular}{|l|r|r|r|r|r|}
\hline
Measure & AAE & CRNN & VAE & ORGAN & Our \\ \hline
$\uparrow$ valid  & 0.60 & 0.351 & 0.001 & 0.108 & \textbf{1.0} \\ %\hline
$\uparrow$ uniq@1K  & 0.36 & \textbf{1.0} & \textbf{1.0} & 0.944 & \textbf{1.0} \\ \hline
$\downarrow$ FCD/Test  & 33.9 & \textbf{6.1} & nan & 33.2 & 35.8 \\ %\hline
$\uparrow$ Frag/Test  & 0.91 & \textbf{0.98} & 0.0 & 0.56 & 0.89 \\ %\hline
$\uparrow$ Scaf/Test  & 0.05 & \textbf{0.26} & nan & 0.0 & 0.0 \\ %\hline
$\uparrow$ SNN/Test  & \textbf{0.47} & 0.37 & 0.09 & 0.27 & 0.22 \\ \hline
$\uparrow$ IntDiv  & 0.69 &\textbf{0.85} & 0.0 & 0.85 & 0.83 \\ %\hline
$\uparrow$ IntDiv2  & 0.66 & \textbf{0.84} & 0.0 & 0.81 & 0.82 \\ %\hline
$\uparrow$ Filters  & \textbf{0.96} & 0.93 & 1.0 & 0.58 & 0.87 \\ \hline
\end{tabular}
\caption{
\newline
Results from the MOSES benchmarks comparing various methods for molecule generation. FCD - ChemNet distance; a measurement for molecule likeness, Frag/Scaf - cosine similarity to graph fragments, SNN - similarity to the graphs in the test set, IntDiv - diversity of generated artifacts, Filters - fraction of graphs that is non toxic or not unstable. The graph coarsened graph grammar approach performs similar to the specialized methods while generating only valid graphs. CharRNN generates very small graphs, AAE generates graphs that are very similar to those in the training set. }
\label{mosestab}
\end{table}

\begin{figure}[ht] \centering \includegraphics[width=1\linewidth]{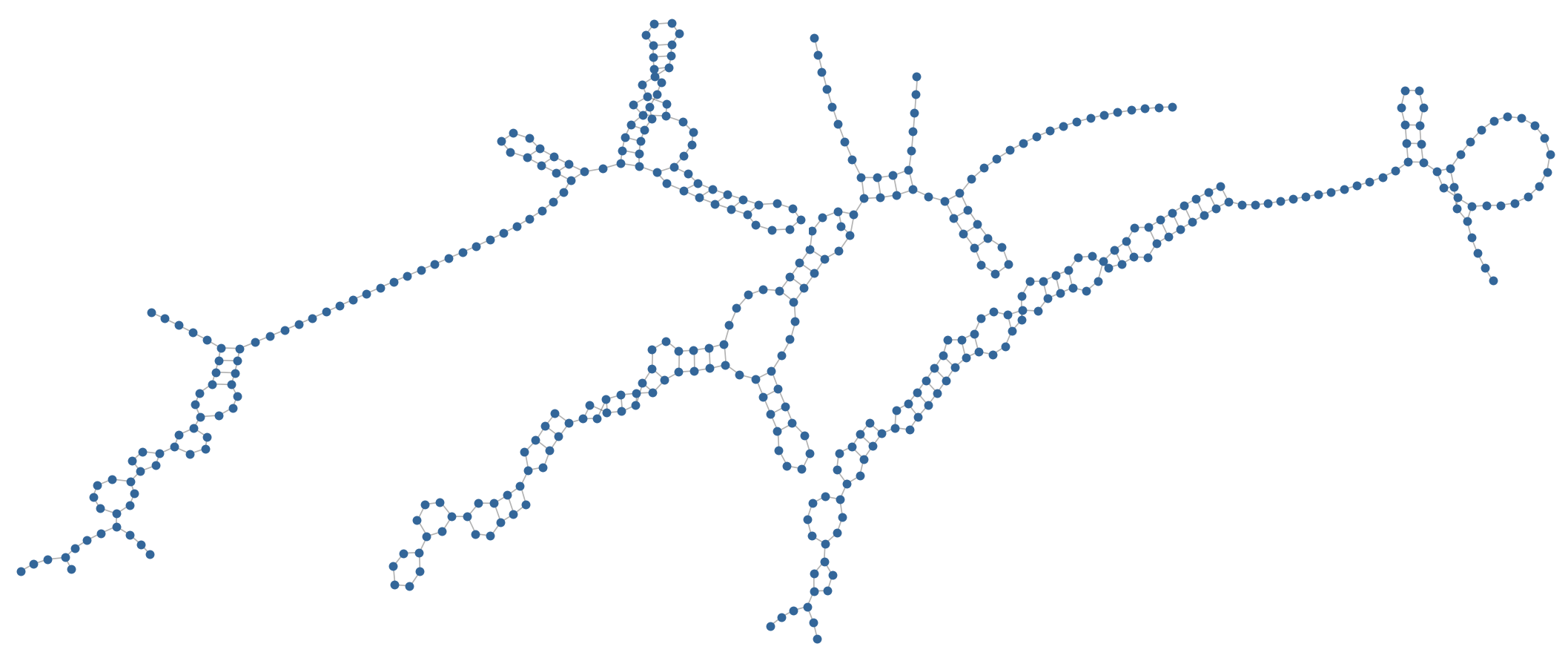}
      \caption{Instances generated by GraphRNN when trained on the seed
      structures of RF01725. The outputs are not structured enough compared to
      Fig.\ref{RF01725} and there is no clear ribose backbone, violating the
      domain constraints of RNA. } \label{graphrnn} \end{figure}

GraphRNN \citep{graphRNN} works with unlabeled graphs and sets out to model long
range dependencies using an auto regressive model. We Trained this method on the
seed sequences of RF01725 whose structure we obtained with the MUSCLE scheme
described previously. In Fig. \ref{graphrnn} we show three graphs generated by
GraphRNN. While the \emph{hairpin} structure is present, in the lower part of
the figure we see cycles and in general more than two \emph{dangling ends}
indicating that the most basic constraint, a continuous ribose backbone, is
violated. Out of 1024 generated graphs only 13 passed the following simple
filters: a) maximal 2 nodes have a degree of 1. b) none has a degree > 3. c) is
connected. d) cycles where exactly 2 nodes have degree 3 and any other vertex
has degree 2, the degree 3 nodes must be directly connected. a) and c) are
necessary for RNA graphs because an nucleotides form a path graph on top of
which H-bonds (edges between non adjacent vertices) are formed. One nucleotide
can only form a single H-bond (filter b). Filter d eliminates "nearly-hairpin"
structures as seen in Fig. \ref{graphrnn}(lower left), were the degree 3 nodes
next to each other, it would form a valid hairpin structure.

% there are this many graphs:  1024
% how many pass the filter: number, percent
% all filters: 13 0.01
% 2x deg1 206 0.20
% max degree 618 0.60
% connected 557 0.54
% cycles 392 0.38

\section{Conclusion}

We have introduced an approach to efficiently model long-range dependencies in graph generation, using a coarsening technique. Our method has been tested on the challenging task of RNA molecule generation, which is particularly difficult due to the large graph sizes, limited training data, and the need to preserve complex global structure constraints. The results are promising, demonstrating that our method can generate novel RNA sequences that are functionally equivalent to the original examples. Additionally, we compared our approach to various specialized systems for generating small molecular graphs using deep neural networks, and in scenarios with limited training data, our method achieved comparable performance. Lastly, we found that these approaches are not well-suited for domains with significant long-range dependencies. In future work, one limitation we aim to address is the reliance on predefined coarsening procedures, which may not be optimal for every domain, especially those with limited domain-specific knowledge. To overcome this, we plan to explore ways to learn coarsening procedures in a data-driven fashion, i.e. leveraging machine learning techniques, we hope to dynamically capture these procedures directly from available data. This approach would not only enhance the adaptability of our method across various domains but also improve its scalability in dealing with highly complex and large graph structures.
In parallel, we will also investigate novel strategies to reduce the computational costs associated with handling long-range dependencies, which is crucial for the efficient application of our method in modeling real-world biological systems.

\section{Funding/Acknowledgements}

This work was supported by the de.NBI Cloud within the German Network for
Bioinformatics Infrastructure (de.NBI) and ELIXIR-DE (Forschungszentrum Jülich
and W-de.NBI-001, W-de.NBI-004, W-de.NBI-008, W-de.NBI-010, W-de.NBI-013,
W-de.NBI-014, W-de.NBI-016, W-de.NBI-022).

\bibliographystyle{unsrtnat}
\bibliography{mybib}  %%% Uncomment this line and comment out the ``thebibliography'' section below to use the external .bib file (using bibtex) .

\begin{thebibliography}{18}
\providecommand{\natexlab}[1]{#1}
\providecommand{\url}[1]{\texttt{#1}}
\expandafter\ifx\csname urlstyle\endcsname\relax
  \providecommand{\doi}[1]{doi: #1}\else
  \providecommand{\doi}{doi: \begingroup \urlstyle{rm}\Url}\fi

\bibitem[Polykovskiy et~al.(2020)Polykovskiy, Zhebrak, Sanchez-Lengeling,
  Golovanov, Tatanov, Belyaev, Kurbanov, Artamonov, Aladinskiy, Veselov,
  Kadurin, Johansson, Chen, Nikolenko, Aspuru-Guzik, and Zhavoronkov]{MOSES}
Daniil Polykovskiy, Alexander Zhebrak, Benjamin Sanchez-Lengeling, Sergey
  Golovanov, Oktai Tatanov, Stanislav Belyaev, Rauf Kurbanov, Aleksey
  Artamonov, Vladimir Aladinskiy, Mark Veselov, Artur Kadurin, Simon Johansson,
  Hongming Chen, Sergey Nikolenko, Alan Aspuru-Guzik, and Alex Zhavoronkov.
\newblock {M}olecular {S}ets ({MOSES}): {A} {B}enchmarking {P}latform for
  {M}olecular {G}eneration {M}odels.
\newblock \emph{Frontiers in Pharmacology}, 2020.

\bibitem[Costa(2017)]{costa16}
Fabrizio Costa.
\newblock Learning an efficient constructive sampler for graphs.
\newblock \emph{Artif. Intell.}, 2017.

\bibitem[Metropolis et~al.(1953)Metropolis, Rosenbluth, Rosenbluth, Teller, and
  Teller]{metropolis1953}
Nicholas Metropolis, Arianna~W Rosenbluth, Marshall~N Rosenbluth, Augusta~H
  Teller, and Edward Teller.
\newblock Equation of state calculations by fast computing machines.
\newblock \emph{The journal of chemical physics}, 21:\penalty0 1087, 1953.

\bibitem[Rozenberg and Ehrig(1999)]{rozenberg1999handbook}
Grzegorz Rozenberg and Hartmut Ehrig.
\newblock \emph{Handbook of graph grammars and computing by graph
  transformation}, volume~1.
\newblock World Scientific, 1999.

\bibitem[Harris(1954)]{harris1954distributional}
Z.S. Harris.
\newblock Distributional structure.
\newblock \emph{WORD}, 10:2--3:\penalty0 146--162, 1954.

\bibitem[Harris(1968)]{harris1968mathematical}
Zellig~Sabbettai Harris.
\newblock Mathematical structures of language.
\newblock 1968.

\bibitem[Clark and Eyraud(2007)]{Clark:2007}
Alexander Clark and R{\'e}mi Eyraud.
\newblock Polynomial identification in the limit of substitutable context-free
  languages.
\newblock \emph{J. Mach. Learn. Res.}, 8:\penalty0 1725--1745, December 2007.

\bibitem[S.~Griffiths-Jones(2003)]{rfam}
M.~Marshall S.~Griffiths-Jones, A.~Bateman.
\newblock Rfam: an rna family database.
\newblock \emph{Nucleic Acids Res.}, 31:\penalty0 439--441, 2003.

\bibitem[Edgar(2004)]{muscle}
Robert~C. Edgar.
\newblock Muscle: multiple sequence alignment with high accuracy and high
  throughput.
\newblock \emph{Nucleic Acids Res.}, 32:\penalty0 1792--1797, 2004.

\bibitem[Stephan H.~Bernhart(2008)]{rnaalifold}
Ivo L.~Hofacker Stephan H.~Bernhart.
\newblock Rnaalifold: improved consensus structure prediction for rna
  alignments.
\newblock \emph{BMC Bioinformatics}, 9:\penalty0 474, 2008.

\bibitem[E.~P.~Nawrocki(2013)]{infernal}
S.~R.~Eddy E.~P.~Nawrocki.
\newblock Infernal 1.1: 100-fold faster rna homology searches.
\newblock \emph{Bioinformatics}, 29:\penalty0 2933--2935, 2013.

\bibitem[Yujia et~al.(2018)Yujia, Oriol, Chris, Razvan, and Peter]{deepGMG}
Li~Yujia, Vinyals Oriol, Dyer Chris, Pascanu Razvan, and Battaglia Peter.
\newblock Learning deep generative models of graphs.
\newblock \emph{CoRR}, abs/1803.03324, 2018.
\newblock URL \url{http://arxiv.org/abs/1803.03324}.

\bibitem[Simonovsky~Martin(2018)]{graphVAE}
Komodakis~Nikos Simonovsky~Martin.
\newblock Graphvae: Towards generation of small graphs using variational
  autoencoders.
\newblock In V{\v{e}}ra K{\r{u}}rkov{\'a}, Yannis Manolopoulos, Barbara Hammer,
  Lazaros Iliadis, and Ilias Maglogiannis, editors, \emph{Artificial Neural
  Networks and Machine Learning -- ICANN 2018}, pages 412--422, Cham, 2018.
  Springer International Publishing.

\bibitem[Segler et~al.(2018)Segler, Kogej, Tyrchan, and Waller]{smilesbasedgen}
Marwin H.~S. Segler, Thierry Kogej, Christian Tyrchan, and Mark~P. Waller.
\newblock Generating focused molecule libraries for drug discovery with
  recurrent neural networks.
\newblock \emph{ACS Central Science}, 4\penalty0 (1):\penalty0 120--131, 2018.
\newblock \doi{10.1021/acscentsci.7b00512}.
\newblock URL \url{https://doi.org/10.1021/acscentsci.7b00512}.

\bibitem[Preuer et~al.(2018)Preuer, Renz, Unterthiner, Hochreiter, and
  Klambauer]{chemnet}
Kristina Preuer, Philipp Renz, Thomas Unterthiner, Sepp Hochreiter, and Günter
  Klambauer.
\newblock Fréchet chemnet distance: A metric for generative models for
  molecules in drug discovery.
\newblock \emph{Journal of Chemical Information and Modeling}, 58\penalty0
  (9):\penalty0 1736--1741, 2018.
\newblock \doi{10.1021/acs.jcim.8b00234}.
\newblock URL \url{https://doi.org/10.1021/acs.jcim.8b00234}.
\newblock PMID: 30118593.

\bibitem[Blaschke et~al.(2017)Blaschke, Olivecrona, Engkvist, Bajorath, and
  Chen]{mosesVAE}
Thomas Blaschke, Marcus Olivecrona, Ola Engkvist, J{\"{u}}rgen Bajorath, and
  Hongming Chen.
\newblock Application of generative autoencoder in de novo molecular design.
\newblock \emph{CoRR}, abs/1711.07839, 2017.
\newblock URL \url{http://arxiv.org/abs/1711.07839}.

\bibitem[Sanchez-Lengeling et~al.(2017)Sanchez-Lengeling, Outeiral, Guimaraes,
  and Aspuru-Guzik]{ORGANIC}
Benjamin Sanchez-Lengeling, Carlos Outeiral, Gabriel~L. Guimaraes, and Alan
  Aspuru-Guzik.
\newblock {Optimizing distributions over molecular space. An
  Objective-Reinforced Generative Adversarial Network for Inverse-design
  Chemistry (ORGANIC)}.
\newblock 8 2017.
\newblock \doi{10.26434/chemrxiv.5309668.v3}.
\newblock URL \url{https://chemrxiv.org/articles/ORGANIC_1_pdf/5309668}.

\bibitem[You et~al.(2018)You, Ying, Ren, Hamilton, and Leskovec]{graphRNN}
Jiaxuan You, Rex Ying, Xiang Ren, William Hamilton, and Jure Leskovec.
\newblock {G}raph{RNN}: Generating realistic graphs with deep auto-regressive
  models.
\newblock In Jennifer Dy and Andreas Krause, editors, \emph{Proceedings of the
  35th International Conference on Machine Learning}, volume~80 of
  \emph{Proceedings of Machine Learning Research}, pages 5708--5717,
  Stockholmsmaessan, Stockholm Sweden, 10--15 Jul 2018. PMLR.
\newblock URL \url{http://proceedings.mlr.press/v80/you18a.html}.

\end{thebibliography}





% \end{thebibliography}

\end{document}